\documentclass{aims}

\usepackage{booktabs}
\usepackage{multirow}
\usepackage{bigstrut}
\usepackage{utfsym}
\usepackage{bm}

\usepackage{txfonts}
\def\typeofarticle{Research article}
\def\currentvolume{19}
\def\currentissue{x}
\def\currentyear{2023}

\def\ppages{xxx--xxx}
\def\DOI{}
\def\Received{xxx xxx}
\def\Revised{xxx xxx}
\def\Accepted{xxx xxx}
\def\Published{xxx xxx}

\numberwithin{equation}{section}

\begin{document}

\title{MIPANet: Optimizing RGB-D Semantic Segmentation through Multi-modal Interaction and Pooling Attention}

\author{%
  Shuai Zhang\affil{1},
  Minghong Xie\affil{1,}\corrauth
}

\shortauthors{the Author(s)}

\address{%
  \addr{\affilnum{1}}{Faculty of Information Engineering and Automation, Kunming University of Science and Technology, Kunming, 650500 Yunnan, PR China}}

\corraddr{minghongxie@163.com.}

\begin{abstract}
Semantic segmentation of RGB-D images involves understanding the appearance and spatial relationships of objects within a scene, which requires careful consideration of various factors. However, in indoor environments, the simple input of RGB and depth images often results in a relatively limited acquisition of semantic and spatial information, leading to suboptimal segmentation outcomes. To address this, we propose the Multi-modal Interaction and Pooling Attention Network (MIPANet), designed to harness the interactive synergy between RGB and depth modalities, optimizing the utilization of complementary information.
Specifically, we incorporate a Multi-modal Interaction Module (MIM) into the deepest layers of the network. This module is engineered to facilitate the fusion of RGB and depth information, allowing for mutual enhancement and correction. Additionally, we introduce a Pooling Attention Module (PAM) at various stages of the encoder to enhance the features extracted by the network. The outputs of the PAMs are selectively integrated into the decoder to improve semantic segmentation performance. Experimental results demonstrate that MIPANet outperforms existing methods on two indoor scene datasets, NYUDv2 and SUN-RGBD, by optimizing the insufficient information interaction between different modalities in RGB-D semantic segmentation. 

\end{abstract}


\maketitle

\section{Introduction}
In recent years, Convolutional Neural Networks (CNN) have been widely used in image semantic segmentation, and more and more high-performance models have gradually replaced the traditional semantic segmentation methods. With the introduction of Fully Convolutional Neural Networks (FCN) \cite{long2015fully, li2022enhancing}, which show great potential in semantic segmentation tasks, many researchers have proposed improved semantic segmentation models based on this way. Nevertheless, semantic segmentation remains a formidable challenge in some indoor environments, given the intricacies such as variations in illumination and mutual occlusion between objects.

\begin{figure}[t]
	\centering
		\centering
\centering
\includegraphics[width=1.0\linewidth]{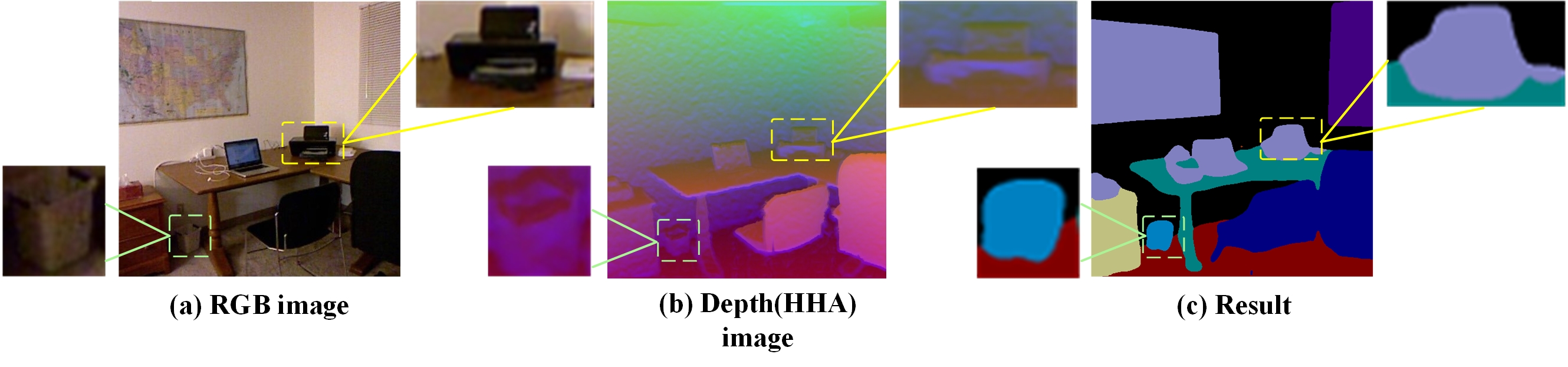}
\caption{Improve segmentation accuracy by leveraging depth features within our MIPANet. The prediction result can accurately distinguish the trash can and printer from the background.\label{fig:example}}
\end{figure}

With the widespread application of depth sensors and depth cameras \cite{zhang2012microsoft}, the research on images is not limited to RGB color images, but the research on RGB-Depth (RGB-D) images containing depth information. RGB images can provide appearance information such as the color and texture of objects, in contrast, depth images can provide three-dimensional geometry information of objects, which is missing in RGB images and is desired for indoor scenes. References \cite{he2017std2p,couprie2013indoor} simply splice RGB features and depth features to form a four-channel input, improving the accuracy of semantic segmentation. Reference \cite{gupta2014learning} convert depth images into three distinct channels (horizontal disparity, height above ground, and angle of surface normals) to obtain the HHA image, then input the RGB features and HHA features into two parallel CNNs to predict the probability maps of two semantic segmentations, respectively, and fuse them in the last layer of the network as the final segmentation result. Though the above methods have achieved good results in the task of RGB-D semantic segmentation, most RGB-D semantic segmentation \cite{park2017rdfnet,jiang2018rednet,eigen2015predicting,wang2014multi} simply merges RGB features and depth features by concatenation or summation. As a result, the information differences between the multimodal cannot be solved effectively, which will generate CNN not to use the complementary information between them fully, resulting in object and background confusion. For example, The printer and trash bin in Fig. \ref{fig:example} (a) are prone to be inaccurately assimilated into the background.

To solve the above problems, we propose an RGB-D semantic segmentation of the Indoor Scene network, MIPANet. Fig. \ref{fig:net} illustrates the overall structure of the network. The network is an encoder-decoder architecture, including two innovative feature fusion modules: The multi-modal Interaction Module(MIM) and the Pooling Attention Module(PAM). This paper integrates the two fusion modules into an encoder-decoder architecture. The encoder is composed of two identical CNN branches, each specifically designed for extracting RGB features and depth features, respectively. In this study, RGB and depth features are extracted and fused incrementally across various network levels, optimizing semantic segmentation results utilizing spatial disparities and semantic interdependencies among multimodal features. In the PAM, we use adaptive averaging instead of global averaging, which approach not only allows for flexible adjustment of the output size but also preserves more spatial information, facilitating enhanced extraction of depth features. In MIM, we obtain two sets of Q,K,V for different modalities and perform calculations using the Q,K from one set and V from the other. This achieves information interaction between the RGB and depth modalities. This paper's main contributions can be summarized as follows:

$\bullet$ We introduce an end-to-end multi-modal fusion network, MIPANet, incorporating multi-modal interaction and pooling attention. This innovative approach optimizes integrating complementary information from RGB and depth features, effectively tackling the challenge posed by insufficient cross-modal feature fusion in RGB-D semantic segmentation.

$\bullet$ We present two cross-modal feature fusion methods. Within the MIM, a cross-modal feature interaction and fusion mechanism were developed. RGB and depth features are collaboratively optimized using attention masks to extract partially detailed features. In addition, PAM integrates intermediate layer features into the decoder, enhancing feature extraction and supporting the decoder in upsampling and recovery.

$\bullet$ Experimental results confirm the effectiveness of our proposed RGB-D semantic segmentation network in accurately handling indoor images in complex scenarios. The model demonstrated superior semantic segmentation performance compared to other methods on the publicly available NYUv2 and SUN RGB-D datasets.

\section{Related Work}
In this section, we provide a comprehensive review of three parts: (1) RGB-D Semantic Segmentation, (2) Attention Mechanism, and (3) Cross-modal Interaction.

\subsection{RGB-D Semantic Segmentation}
With the widespread application of depth sensors and depth cameras in the field of depth estimation \cite{liu2015deep,eigen2015predicting,hu2023bag,hu2023robust}, people can obtain the depth information of the scene more conveniently, and the research on the image is no longer limited to a single RGB image. RGB-D semantic segmentation task is to efficiently integrate RGB features and depth features to improve segmentation accuracy, especially in some indoor scenes. Couprie et al. \cite{he2017std2p} proposed an early fusion approach, which simply concatenates an image's RGB and depth channels as a four-channel input to the convolutional neural network. Wang et al. \cite{gupta2014learning} separately input RGB features and HHA features into two CNNs for prediction and perform fusion in the final stage of the network, and \cite{hazirbas2017fusenet} introduced an encoding-decoding network, employing a dual-branch RGB encoder to extract features separately from RGB images and depth images. The studies mentioned above employed equal-weight concatenation or summation operations to fuse RGB and depth features without fully leveraging the complementary information between different modalities. In recent years, some research has proposed more effective strategies for RGB-D feature fusion. Hu et al. \cite{hu2019acnet} utilised a three-branch encoder that includes RGB, Depth, and Fusion branches, efficiently collecting features without breaking the original RGB and deep inference branches. Seichter et al. \cite{seichter2021efficient} have presented an efficient RGB-D segmentation approach, characterised by two enhanced ResNet-based encoders utilising an attention-based fusion for incorporating depth information. However, these methods did not fully exploit the differential information between the two modalities and the intermediate-level features extracted by the convolutional network.

\subsection{Attention Mechanism}
In recent years, attention \cite{vaswani2017attention, shen2022hsgnet, fu2019dual, shen2021efficient, shen2022hsgm, woo2018cbam} has been widely used in computer vision and other fields. Vaswani et al. \cite{vaswani2017attention} proposed the self-attention mechanism, which has had a profound impact on the design of the deep learning model. Fu et al. \cite{fu2019dual} proposed DANet, which can adaptively integrate local features and their global dependencies. Wang et al. \cite{wang2017residual} utilised spatial attention in an image classification model. Through the backpropagation of a convolutional neural network, they adaptively learned spatial attention masks, allowing the model to focus on the significant regions of the image. SENet \cite{hu2018squeeze} has proposed channel attention, which adaptively learns the importance of each feature channel through a neural network. Woo et al. \cite{woo2018cbam} incorporates two attention modules that concurrently capture channel-wise and spatial relationships. ECA-Net \cite{wang2020eca} introduces a straightforward and efficient "local" channel attention mechanism to minimize computational overhead. MFC \cite{qiao2022novel}introduced a multi-frequency domain attention module to capture information across different frequency domains. Similarly, CAMNet \cite{li2022enhancing} proposed a contrastive attention module designed to amplify local saliency. Building upon this foundation, Huang et al. \cite{huang2019ccnet} proposed a cross-attention module that consolidates contextual information both horizontally and vertically, which can gather contextual information from all pixels. These methods have demonstrated significant potential in single-mode feature extraction. To effectively leverage the complementary information between different modalities, this paper introduces a Pooling Attention module that learns the differential information between two distinct modalities and fully exploits the intermediate-level features in the convolutional network and long-range semantic dependencies between modalities.

\subsection{Cross-modal Interaction}
With the development of sensor technology, different types of sensors can provide a variety of modal information for semantic segmentation tasks to achieve information interaction \cite{ha2017mfnet, shen2023triplet, zhang2021abmdrnet, shen2023pbsl, sun2019rtfnet}  between RGB mode and other modes.  The interaction between RGB and infrared modalities enhanced the effectiveness of semantic segmentation in RGB-T scenarios. Xiang et al. \cite{xiang2021polarization} used a single-shot polarization sensor to build the first RGB-P dataset, incorporated polarization sensing to obtain supplementary information, and improved the accuracy of segmentation for many categories, especially those with polarization characteristics, such as glass. HPGN \cite{shen2021exploring} proposes a novel pyramid graph network targeting features, which is closely connected behind the backbone network to explore multi-scale spatial structural features. GiT \cite{shen2023git}  proposes a structure where graphs and transformers interact constantly, enabling close collaboration between global and local features for vehicle re-identification. Zhuang et al. \cite{zhuang2021perception} propose a network consisting of a two-streams (LiDAR stream and camera stream), which extract features from two modes respectively to realize information interaction between RGB and LIDAR modes. Improving the result of semantic segmentation by information interaction between different modes and RGB mode is feasible.

\section{Methods}

\subsection{Overview} 
\begin{figure}[t]
\centering
\includegraphics[width=1.0\linewidth]{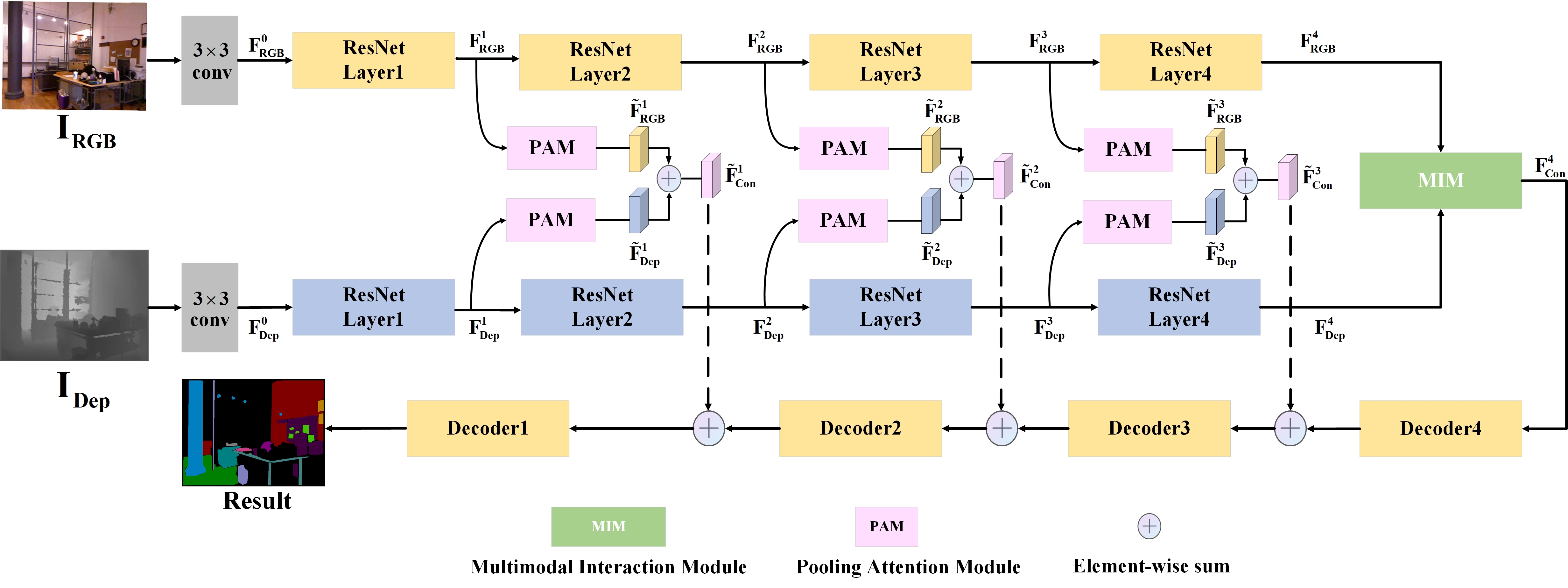}
\caption{Multi-modal Interaction And Pooling Attention (MIPA) Network architecture. Each PAM at different network levels generates two weight-unshared features: RGB features denoted as $ \tilde{\bm F}_{RGB}^{n}$ and depth features denoted as $\tilde{\bm F}_{Dep}^{n}$. Following an Element-wise sum, we obtain $\tilde{\bm F}_{Con}^{n}$, where n denotes the network level. MIM receives RGB and depth features from the ResNetLayer4 and integrates the fusion result ${\bm F}_{Con}^{4}$ into the decoder.\label{fig:net}}
\end{figure}

Fig. \ref{fig:net} depicts the overall structure of the network. The architecture follows an encoder-decoder design, employing skip connections to facilitate information flow between encoding and decoding layers. The encoder comprises a dual-branch convolutional network, with each branch respective to extracting RGB features and depth features. We utilize two pre-trained ResNet50 models as the backbone, which exclude the final global average pooling layer and fully connected layers. Subsequently, a decoder is employed to upsample the features, progressively restoring image resolution incrementally.

Given a RGB image $ {I}_{RGB} \in {\mathbb{R}}^{h\times w \times 3}$, and a Depth image $ {I}_{Dep} \in {\mathbb{R}}^{h\times w \times 1}$, $3 \times 3$ convolution is used to extract them shallow features ${\bm F}_{RGB}^{0}$ and ${\bm F}_{Dep}^{0}$, which can be expressed as:
\begin{equation}
    \label{eq1}
    {\bm F}_{RGB}^{0} = {Conv}_{3\times3}({I}_{RGB})
\end{equation}

\begin{equation}
    \label{eq2}
    {\bm F}_{Dep}^{0} = {Conv}_{3\times3}({I}_{Dep})
\end{equation}
where ${Conv}_{3\times3}$ denotes $3 \times 3$ convolution.

The network mainly consists of a four-layer encoder-decoder and introduces two feature fusion modules: MIM and the PAM. Each layer of the encoder consistes of a ResNetLayer. After ${\bm F}_{i}^{0}$ passing through the ResNetLayer, ${\bm F}_{i}^{n}$ is obtained, the n-th layer of the encoder can be expressed as:
\begin{equation}
    \label{eq3}
    {\bm F}_{i}^{n} = H_{i}^{n}({\bm F}_{i}^{n-1})
\end{equation}
where ${H}_{i}^{n}$ (n = 1, 2, 3, 4) represents the n-th ResNetLayer, $ i\in \{RGB,Depth\}$ denotes the RGB feature or Depth feature. Specifically, the first three multi-level RGB features (ResNetLayer1-ResNetLayer3) and depth features (ResNetLayer1-ResNetLayer3) of the ResNet encoder are fed into the PAM module. Pooled attention weighting operations are performed on the RGB features and depth features separately to obtain $\tilde{\bm F}_{RGB}^{n}$ and $\tilde{\bm F}_{Dep}^{n}$, where n = 1, 2, 3. Subsequently, the two features are combined by element-wise addition to obtain $\tilde{\bm F}_{Con}^{n}$, containing rich spatial location information. Furthermore, the final RGB and depth features from the ResNetLayer4 encoder are fed into the MIM module to capture complementary information within these two modalities. The output features of the MIM module are then fed into the decoder, where each upsampling layer consists of two 3 $\times$ 3 convolutional layers. These layers are followed by batch normalization (BN) and ReLU activation, with each upsampling layer doubling the feature spatial dimensions while halving the number of channels.

\subsubsection{Pooling Attention Module}
\begin{figure}[t]
	\centering
		\centering
\centering
\includegraphics[width=1.0\linewidth]{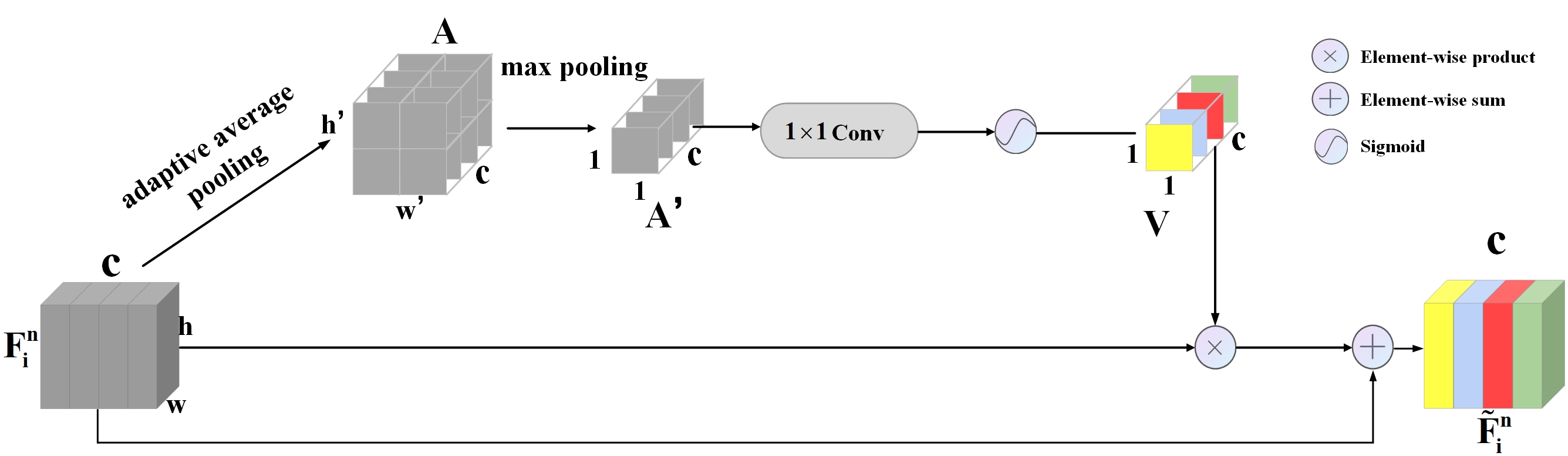}
\caption{The details of the Pooling Attention Module. After a two-step pooling operation, we obtain the pooling result $\bm A^\prime$. Subsequently, through a 1 $\times$ 1 convolution and sigmoid activation function, constrain the value of weight vector $\bm V$ (e.g., yellow) between 0 and 1. The output feature $\tilde{\bm F}_{i}^{n}$ is obtained by taking the weighted sum of the input feature ${\bm F}_{i}^{n}$.\label{fig:pam}}
\end{figure}

Within the low-level features extracted by the convolutional neural network, we capture the fundamental attributes of the input image. These low-level features are critical in modelling the image's foundational characteristics. However, they lack semantic information from the high-level neural network, such as object shapes and categories. At the same time, during the upsampling process in the decoding layer, there is a risk of losing certain semantic information as the image resolution increases. We introduce the Pooling Attention Module (PAM) to address this issue. The PAM module enhances the representation of these features by using an attention mechanism to focus on critical areas in the low-level feature map. In the decoding layer, we integrate the PAM module's output with the upsampling layer's input, effectively compensating for information loss during the upsampling process. This strategy improves the accuracy of segmentation results and efficiently maintains the integrity of semantic information, as shown in Fig. \ref{fig:pam}.

The input featre $\bm {\bm F}_{i}^{n}\in{\mathbb{R}}^{h\times w \times c}$ where $ i\in \{RGB,Depth\}$ denotes the RGB feature or Depth feature passes through adaptive average pooling to reduce the feature map to a smaller dimension:

\begin{equation}
    \label{eq4}
    \bm A = {H}_{ada}({\bm F}_{i}^{n})
\end{equation}
where $\bm A \in {\mathbb{R}}^{{h}^{\prime} \times {w}^{\prime} \times c}$ represents the feature map that has been resized by adaptive averaging pooling, ${H}_{ada}$ denotes the adaptive average pooling operation. ${h}^{\prime}$,${w}^{\prime}$ represent the height and width of the output feature map, which we set ${h}^{\prime} = 2 $ and ${w}^{\prime} = 2 $. Then we get the output features ${\bm A}^{\prime}$ by max pooling the features after dimensionality reduction:
\begin{equation}
    \label{eq5}
    {\bm A}^{\prime} = {H}_{max}(\bm A)
\end{equation}
where $ {\bm A}^{\prime} \in {\mathbb{R}}^{1\times 1 \times c}$ represents the pooling result and then ${\bm A}^{\prime}$ undergoes a 1 $\times$ 1 convolution and then activation with the sigmoid function, getting a weight vector $\bm V$ $\in {\mathbb{R}}^{1 \times 1\times c}$ value between 0 and 1. ${H}_{max}$ denotes the max pooling operation. Finally, we perform an Element-wise product for ${\bm F}_{i}^{n}$ and $\bm V$, and the result $ \tilde{\bm F}_{i}^{n} $ can be expressed as:
\begin{equation}
    \label{eq6}
    {\bm V} = Sigmoid(\varPhi (\bm {A}^{\prime}))
\end{equation}

\begin{equation}
    \label{eq7}
    \tilde{\bm F}_{i}^{n} = {\bm F}_{i}^{n} + ({\bm F}_{i}^{n} \otimes \bm V)
\end{equation}
where $\otimes$ denotes the Element-wise product, $\varPhi$ denotes 1 $\times$ 1 convolution, and feature maps $\tilde{\bm F}_{i}^{n}$ represent the output feature $\tilde{\bm F}_{RGB}^{n}$ or $\tilde{\bm F}_{Dep}^{n}$ in Fig. \ref{fig:net}. We employ two-step pooling operation instead of conventional global average pooling. Firstly, the input features ${\bm F}_{i}^{n}$ pass through adaptive average pooling to obtain the middle feature $\bm A$ with a specified output size. Then, $\bm A$ undergoes max pooling to yield the final result $\bm A^{\prime}$. This modification makes the network pay more attention to local regions in the image, such as objects near the background in the scene. Meanwhile, adapt average pooling can enhance the module's flexibility, accommodating diverse input feature map dimensions and fully retaining spatial position information in depth features; the visualization results Fig. \ref{fig:view} show the module's effectiveness. The final output $\tilde{\bm F}_{Con}^{n}$ of the PAM in Fig. \ref{fig:net}:
\begin{equation}
    \label{eq8}
    \tilde{\bm F}_{Con}^{n} = \tilde{\bm F}_{RGB}^{n} + \tilde{\bm F}_{Dep}^{n}
\end{equation}
During the upsampling process, $\tilde{\bm F}_{Con}^{n}$ (n = 1, 2, 3) will play a role in the three-level decoder (decoder1-decoder3).

\subsection{Multi-modal Interaction Module}  

When adjacent objects in an image share similar appearances, distinguishing their categories becomes challenging. Factors such as lighting variations and object occlusion, especially in the corners, can lead to their blending with the background. This complexity makes it difficult to precisely identify object edges, leading to misclassification of the object as part of the background. Depth information remains unaffected by lighting conditions and can accurately differentiate between objects and the background based on depth values. Therefore, we designed the MIM module to supplement RGB information with Depth features. Meanwhile, it utilizes RGB features to strengthen the correlation between RGB and depth features.

\begin{figure}[t]
	\centering
		\centering
\centering
\includegraphics[width=1.0\linewidth]{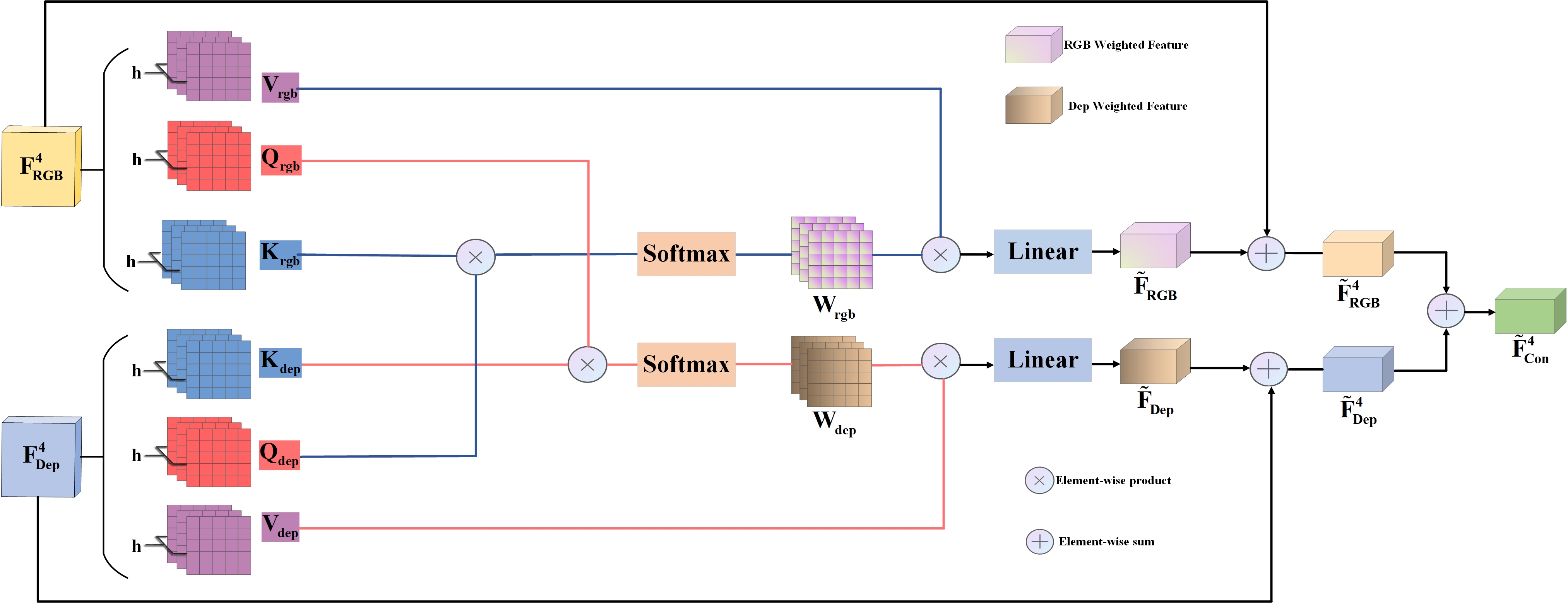}
\caption{Multi-modal Interaction Module. The RGB feature and the depth feature undergo linear transformations to generate two sets of Q,K,V (e.g., blue line) for multi-head attention, where h denotes the number of attention heads set to 8. The weighted summation of input features ${\bm F}_{RGB}^{4}$ and ${\bm F}_{Dep}^{4}$ yields $\tilde{\bm F}_{RGB}^{4}$ and $\tilde{\bm F}_{Dep}^{4}$, which are then element-wise added to obtain the output result $\tilde{\bm F}_{Con}^{4}$. \label{fig:MIM}}
\end{figure}

The Multi-modal Interaction Module achieves dual-mode feature fusion, as depicted in Fig. \ref{fig:MIM}. Here, ${\bm F}_{RGB}^{4}\in {\mathbb{R}}^{h\times w\times c}$ and ${\bm F}_{Dep}^{4}\in {\mathbb{R}}^{h\times w\times c}$ correspond to the RGB feature and depth feature from the ResNetLayer4. The feature channels are denoted as 'c', and their spatial dimensions are h $\times$ w. First, the two feature maps are linearly mapped to generate multi-head query(Q), key(K), and value(V) vectors. Here, 'rgb' and 'dep' represent the RGB and depth features. These linear mappings are accomplished via fully connected layers, where each attentional head possesses its unique weight matrix. For each attention head, We calculate the dot product between two sets of Q and K and then normalize the results to a range between 0 and 1 using the softmax function to get the transmembrane state attention mask ${\bm W}_{rgb}$ and ${\bm W}_{dep}$:

\begin{equation}
    \label{eq9}
    {\bm W}_{rgb} = Softmax({\bm Q}_{rgb}{\bm K}_{dep}^T / sqrt(d\_k))
\end{equation}
\begin{equation}
    \label{eq10}
    {\bm W}_{dep} = Softmax({\bm Q}_{dep}{\bm K}_{rgb}^T / sqrt(d\_k))
\end{equation}
where ${\bm W}_{rgb}$ and ${\bm W}_{dep}$ represent the RGB attention mask and the Depth attention mask, and d\_k is the dimension of the vector. Then we calculate the RGB Weighted Feature $\tilde{\bm F}_{RGB} $ and the Dep Weighted Feature $\tilde{\bm F}_{Dep} $. We obtain the final output features $\tilde{\bm F}_{RGB}^{4}$ and $\tilde{\bm F}_{Dep}^{4}$ through the use of a residual connection:

\begin{equation}
    \label{eq11}
    \tilde{\bm F}_{RGB}= {\bm W}_{rgb} \otimes {\bm V}_{rgb}
\end{equation}
\begin{equation}
    \label{eq12}
    \tilde{\bm F}_{RGB}^{4} = \tilde{\bm F}_{RGB} + {\bm F}_{RGB}^{4}
\end{equation}
where $\tilde{\bm F}_{RGB}$ represent the RGB Weighted Feature,${\bm V}_{rgb}$ represent the value vector from the RGB feature, multiplying with weight matrix ${\bm W}_{rgb}$. $\tilde{\bm F}_{RGB}^{4}$ represents the RGB feature after the fusion with Depth. Likewise:
\begin{equation}
    \label{eq13}
    \tilde{\bm F}_{Dep}= {\bm W}_{dep} \otimes {\bm V}_{dep}
\end{equation}
\begin{equation}
    \label{eq14}
    \tilde{\bm F}_{Dep}^{4} = \tilde{\bm F}_{Dep} + {\bm F}_{Dep}^{4}
\end{equation}
where $\tilde{\bm F}_{Dep}$ represent the Depth Weighted Feature, ${\bm V}_{dep}$ represent the value vector from the Depth feature, multiplying with weight matrix ${\bm W}_{dep}$. $\tilde{\bm F}_{Dep}^{4}$ represents the Depth feature after the fusion with RGB, $\otimes$ represents the Element-wise product. Finally, we can obtain the MIM output through Element-wise sum, which can be formulated as:
\begin{equation}
    \tilde{\bm F}_{Con}^{4} = \tilde{\bm F}_{RGB}^{4} + \tilde{\bm F}_{Dep}^{4}
\end{equation}

\subsection{Loss Function}

In this paper, the network performs supervised learning on four different levels of decoding features. We employ nearest-neighbor interpolation to reduce the resolution of semantic labels.  Additionally, 1 $\times$ 1 convolutions and Softmax functions are utilized to compute the classification probability for each pixel within the output features from the four upsample layers, respectively. The loss function $\mathcal{L}_{i}$ of layer i is the pixel-level cross entropy loss: 
\begin{equation}
    \mathcal{L}_{i} = -\frac{1}{{N}_{i}}\displaystyle\sum_{\forall p,q}Y(p,q)\log_{}{({Y}^{\prime}(p,q))}
\end{equation}
where ${N}_{i}$ denotes the number of pixels in layer i, p,q is the pixel position, ${Y}^{\prime}$ is the classification probability of the output, and $Y$ is the label category. The final loss function $\mathcal {L}$ of the network is obtained by summing the pixel-level loss functions of the four decoding layers:
\begin{equation}
    \mathcal{L}= \displaystyle\sum_{i=1}^{4}\mathcal{L}_{i}
\end{equation}
By optimizing the above loss function, the network can get the final segmentation result after one training.

\section{Experimental result and analysis}

\subsection{Datasets and Evaluation Measures}
{\bfseries NYU-Depth V2 dataset \cite{silberman2012indoor}} is a widely used indoor scene understanding dataset for computer vision and deep learning research. It is an aggregation of video sequences from various indoor scenes recorded by RGB-D cameras from the Microsoft Kinect and is an updated version of the NYU-Depth dataset published by Nathan Silberman and Rob Fergus in 2011. It contains 1449 RGBD images, depth images, and semantic tags in the indoor environment. The dataset includes different indoor scenes, scene types, and unlabeled frames, and each object can be represented by a class and an instance number.

{\bfseries SUN RGB-D dataset \cite{song2015sun}} contains image samples from multiple scenes, covering various indoor scenes such as offices, bedrooms, and living rooms. It has 37 categories and contains 10335 RGBD images with pixel-level annotations, of which 5285 are used as training images and 5050 are used as test images. This special dataset is captured by four different sensors: Intel RealSence, Asus Xtion, Kinect v1, and v2. Besides, this densely annotated dataset includes 146,617 2D polygons, 64,595 3D bounding boxes with accurate object orientations, and a 3D room layout as well as an imaged-based scene category.We evaluate the results using two standard metrics, Pixel Accuracy (Pixel Acc), and Mean Intersection Over Union (mIoU).

{\bfseries mIoU:} Intersection over Union is a measure of semantic segmentation, where the intersection over Union ratio of a class is the ratio of the intersection over Union of its true labels and predicted values, while mIoU is the average intersection over Union ratio of each class in the dataset.

\begin{equation}
\label{eq15}
	mIoU= \dfrac{1}{k+1} \sum_{i=0}^{k}\frac{{p}_{ii}}{\sum_{j=0}^{k}{p}_{ij}+\sum_{j=0}^{k}{p}_{ji}-{p}_{ii}}.
\end{equation}
where ${p}_{ij}$ represents the predict i as j, and ${p}_{ji}$ represents the predict j as i, ${p}_{ii}$ means to predict the correct value, k represents the number of categories.

{\bfseries Acc:} Pixel accuracy refers to pixel accuracy, which is the simplest metric that represents the proportion of correctly labelled pixels to the total number of pixels.

\begin{equation}
\label{eq16}
	PA=\dfrac{\sum_{i=0}^{k}{p}_{ii}}{\sum_{i=0}^{k}\sum_{j=0}^{k}{p}_{ij}}.
\end{equation}
where ${p}_{ii}$ means to predict the correct value, and ${p}_{ij}$ means to predict i to j.k represents the number of categories.
 
\subsection{Implementation Details}

We implemented and trained our proposed network model using the PyTorch framework. To enhance the diversity of the training data, we applied random scaling and mirroring. Subsequently, all RGB and depth images were resized to $480\times480$ for network inputs, and semantic labels were adjusted to sizes of $480\times480$, $240\times240$, $120\times120$, and $60\times60$ for deep supervision training. As the backbone for our encoder, we utilized a pre-trained ResNet50 \cite{he2016deep} from the ImageNet classification dataset \cite{russakovsky2015imagenet}. To refine the network structure, following \cite{fu2022bag, shen2021competitive, shen2022competitive},  we adjust it by replacing the $7\times7$ convolution in the input stem with three consecutive $3\times3$ convolutions. The training was conducted on an NVIDIA GeForce GTX 3090 GPU using stochastic gradient descent optimization. Parameters were set with a batch size of 6, an initial learning rate of 0.003, 500 epochs, and momentum and weight decay values of 0.9 and 0.0005, respectively.

\subsection{Quantitative Results on NYUv2 and SUN RGB-D}
Firstly, we compare the proposed method against existing approaches using the NYUv2 dataset.

\begin{table}[!htb]
  \centering
  \caption{MIPANet compared to the state-of-the-art methods on the NYUDv2 dataset.}
    \begin{tabular}{lllllllllllll}
    \toprule
    \toprule
    Model &       &       & Method &       &        Backbone &       &       & mIoU(\%)  &       &        Pix.Acc(\%) \\
    \midrule
    ResNet34 &       &       & IEMNet\cite{xu2023interactive}  &       &        Res34NBt1D &       &       & 51.3  &       &       76.8 \\
    \midrule
    &       &       &       &       &       &       &       &       &       &       &         \\
    ResNet18 &       &       & ESANet\cite{seichter2021efficient} &       &        2 $\times$ R18 &       &       & 48.2  &       &        - \\
    \midrule
          &       &       & RDFNet\cite{park2017rdfnet} &       &        2 $\times$  R50 &       &       & 47.7  &       &       74.8 \\
          &       &       &       &       &       &       &       &       &       &       &         \\
          &       &       & ACNet\cite{hu2019acnet} &       &        3 $\times$  R50 &       &       & 48.3  &       &        - \\
          &       &       &       &       &       &       &       &       &       &       &         \\
          &       &       & SA-Gate\cite{chen2020bi} &       &        2 $\times$  R50 &       &       & 50.4  &       &        - \\
    ResNet50 &       &       &       &       &       &       &       &       &       &       &         \\
          &       &       & ESANet &       &        2 $\times$  R50 &       &       & 50.5  &       &        - \\
          &       &       &       &       &       &       &       &       &       &       &         \\
          &       &       & DynMM\cite{xue2023dynamic} &       &        R50   &       &       & 51.0    &       &       - \\
          &       &       &       &       &       &       &       &       &       &       &         \\
          &       &       & RedNet\cite{jiang2018rednet} &       &        2 $\times$  R50 &       &       & 47.2  &       &        - \\
    \midrule
          &       &       & SGNet\cite{chen2021spatial} &       &        R101  &       &       & 49.6  &       &        75.6 \\
          &       &       &       &       &       &       &       &       &       &       &         \\
    ResNet101 &       &       & RDFNet &       &        2 $\times$  R101 &       &       & 49.1  &       &       75.6 \\
          &       &       &       &       &       &       &       &       &       &       &         \\
          &       &       & ShapeConv\cite{cao2021shapeconv} &       &        R101  &       &       & 51.3  &       &        76.4 \\
    \midrule
          &       &       & Baseline &       &        2 $\times$  R50 &       &       & 47.4  &       &        75.1 \\
    ResNet50 &       &       &       &       &       &       &       &       &       &       &         \\
          &       &       & Ours(MIPA) &       &        2 $\times$  R50 &       &       & \textbf{51.9} \textcolor{red}{(+ 4.5\%)}&       &        \textbf{77.2} \textcolor{red}{(+ 2.1\%)} \\
    \bottomrule
    \end{tabular}%
  \label{tab:1}%
\end{table}

\begin{table}[t]
  \centering
  \caption{MIPANet compared to the state-of-the-art methods on the SUN RGB-D dataset.}
    \begin{tabular}{lllllllllllll}
    \toprule
    \toprule
    Model &       &       & Method &       &        Backbone &       &       & mIoU(\%)  &       &        Pix.Acc(\%) \\
    \midrule
          &       &       &       &       &       &       &       &       &       &       &         \\
    ResNet34 &       &       & EMSANet\cite{seichter2022efficient} &       &        2 $\times$  R34 &       &       & 48.5  &       &        - \\
          &       &       &       &       &       &       &       &       &       &       &         \\
         &       &       & IEMNet\cite{xu2023interactive} &       &        Res34NBt1D &       &       & 48.3  &       &        81.9 \\
          
    \midrule
          &       &       & ACNet\cite{hu2019acnet} &       &        3 $\times$  R50 &       &       & 48.1  &       &        - \\
          &       &       &       &       &       &       &       &       &       &       &         \\
    ResNet50 &       &       & ESANet\cite{seichter2021efficient} &       &        2 $\times$  R50 &       &       & 48.3  &       &        - \\
          &       &       &       &       &       &       &       &       &       &       &         \\
          &       &       & RedNet\cite{jiang2018rednet} &       &        2 $\times$  R50 &       &       & 47.8  &       &        81.3 \\
    \midrule
          &       &       & SGNet\cite{chen2021spatial} &       &       R101  &       &       & 47.1  &       &        81.0 \\
          &       &       &       &       &       &       &       &       &       &       &         \\
          &       &       & CANet\cite{tang2021attention} &       &        R101  &       &       & 48.3  &       &        82.0 \\
    ResNet101 &       &       &       &       &       &       &       &       &       &       &         \\
          &       &       & CGBNet\cite{ding2020semantic} &       &        R101  &       &       & 48.2  &       &        82.3 \\
          &       &       &       &       &       &       &       &       &       &       &         \\
          &       &       & ShapeConv\cite{cao2021shapeconv} &       &        R101  &       &       & 48.6  &       &        82.2 \\
    \midrule
          &       &       &       &       &       &       &       &       &       &       &         \\
    ResNet152 &       &       & RDFNet\cite{park2017rdfnet} &       &        2 $\times$  R152 &       &       & 47.7  &       &        81.5 \\
          &       &       &       &       &       &       &       &       &       &       &         \\
          &       &       & Baseline &       &        2 $\times$  R50 &       &       & 45.5  &       &        81.1 \\
    ResNet50 &       &       &       &       &       &       &       &       &       &       &         \\
          &       &       & Ours(MIPA) &       &        2 $\times$  R50 &       &       & \textbf{48.8} \textcolor{red}{(+ 3.3\%)} &       &        \textbf{82.3} \textcolor{red}{(+ 1.2\%)} \\
    \bottomrule
    \end{tabular}%
  \label{tab:2}%
\end{table}%
Table \ref{tab:1} illustrates our superior performance regarding mIoU and Acc metrics compared to other methods. Specifically, with ResNet50 serving as the encoder in our network, the pixel accuracy and average intersection-over-union (mIoU) for semantic segmentation on the NYUv2 test set reached 77.2\% and 51.9\%. For example, contrasting our method with RDFNet, which also employs ResNet50, our approach showcased a notable improvement of 2.4\% in accuracy (Acc) and 3.2\% in mean IoU (mIoU). This underscores a significant enhancement in segmentation accuracy achieved by our MIPANet, leveraging the identical ResNet50 architecture. Compared to SGNet, which utilizes ResNet101, our model demonstrates an improvement of 1.6\% and 2.3\% in Acc and mIoU, respectively. Notably, our ResNet50 outperforms ResNet101, showcasing the effectiveness of our carefully designed network structure and the multi-modal feature fusion module. These improvements in segmentation results are achieved without the need for complex networks, leading to reduced training time. Here, "R" represents ResNet, and the symbol '-' signifies that the comparison evaluated no accuracy metrics. We further compared different network structures across various methods, explicitly noting that ESANet incorporates two ResNet18s as the backbone, while ACNet utilizes three ResNet50 as the backbone.

Then, we comprehensively compared our proposed algorithm with existing methods on the SUN RGB-D dataset. As depicted in Table \ref{tab:2}, our approach consistently achieves higher mIoU scores on the SUN RGB-D dataset than all other methods.   For instance, MIPANet outperforms SGNet, exhibiting an improvement of 1.3\% and 1.7\% in Acc and mIoU, respectively. This observation underscores our module's ability to maintain superior segmentation accuracy, even when dealing with the extensive SUN RGB-D dataset. For different backbone architectures, ResNet101 generally demonstrates better performance than ResNet50, while ResNet50, in turn, outperforms ResNet18. We opted for ResNet50 as our backbone to achieve commendable performance with reduced training time compared to ResNet101.  Notably, our method exhibits an increase of 4.5\% and 2.1\% in mIoU and Acc on both datasets, respectively, compared to the baseline, as highlighted in the red section of the tables.

\subsection{Visualization results on NYUv2}

To visually highlight the advancements made by our method in the realm of RGB-D semantic segmentation, we provide visualization results of the network on the NYUv2 dataset. 
\begin{figure}[!htb]
	\centering
\includegraphics[width=1.0\linewidth]{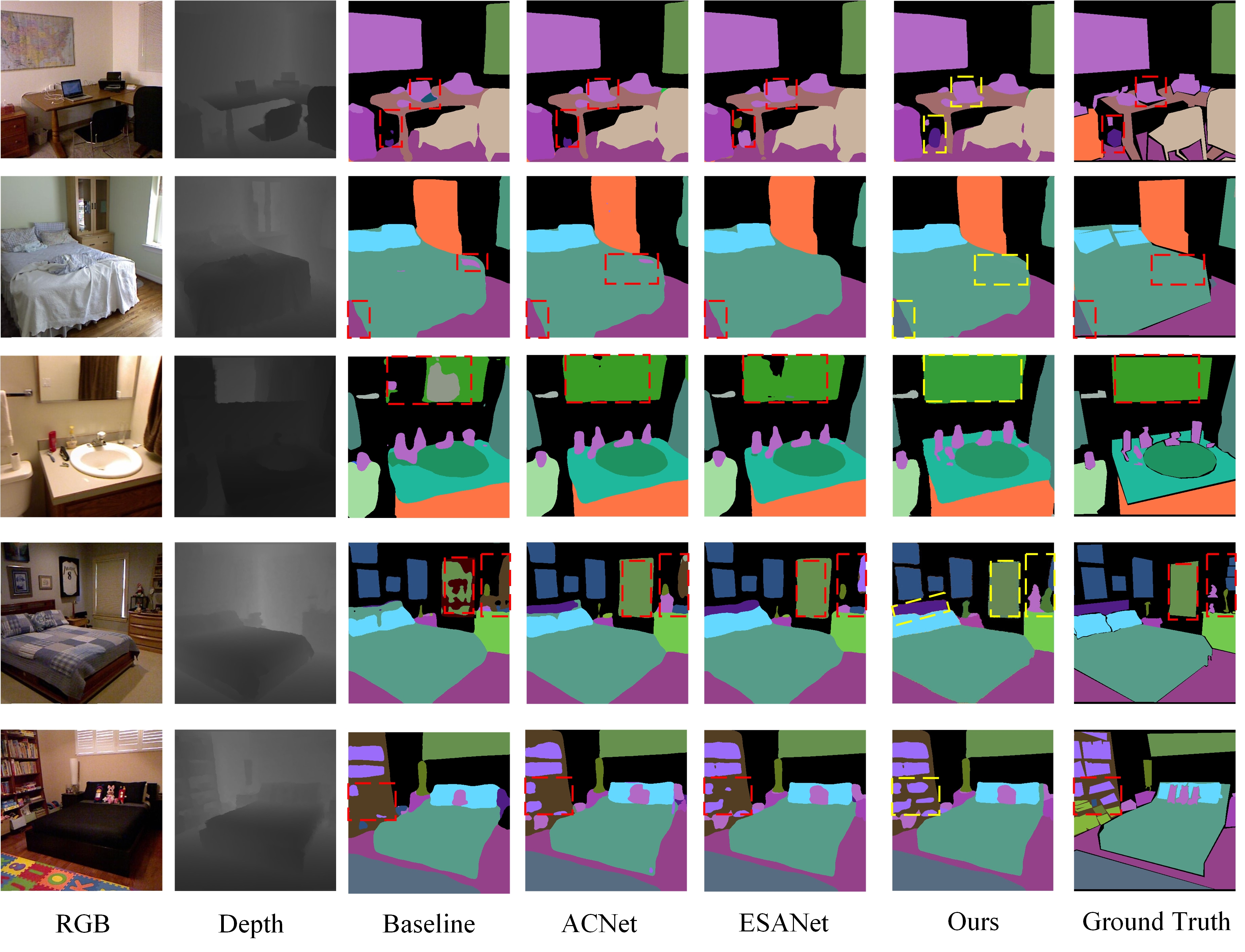}
\caption{Visual result of MIPANet on NYUv2 dataset. The optimization effect is particularly notable within the red dotted box.\label{fig:view}}
\end{figure}
Compared to the baseline, our method has significantly improved segmentation results. Notably, the dashed box in the figure showcases our network enriched with depth information accurately distinguishes objects from the background. For instance, in the visualization results of the fourth image, the baseline erroneously categorizes the mirror on the wall as part of the background, in the visualization results of the second image, the ACNet and the ESANet mistook the carpet for a part of the floor. In contrast, leveraging depth information, our network discerns the distinct distance information of the mirror from the background, leading to a correct classification of the mirror. Fig. \ref{fig:view} illustrates the visualization results of the proposed algorithm on the NYUv2 dataset. From left to right, the columns depict the RGB image, the Depth image, the baseline model results with ResNet50 backbone, ACNet, ESANet, MIPANet (Ours), and Ground Truth. The algorithm presented in this paper has achieved precise segmentation outcomes in diverse and intricate indoor scenes. Moreover, it excels in segmenting challenging objects like "carpets" and "books" while delivering finer-edge segmentation results.

\subsection{Ablation Study on PAM and MIM on NYUv2}

\begin{figure}[!htb]
	\centering
\includegraphics[width=1.0\linewidth]{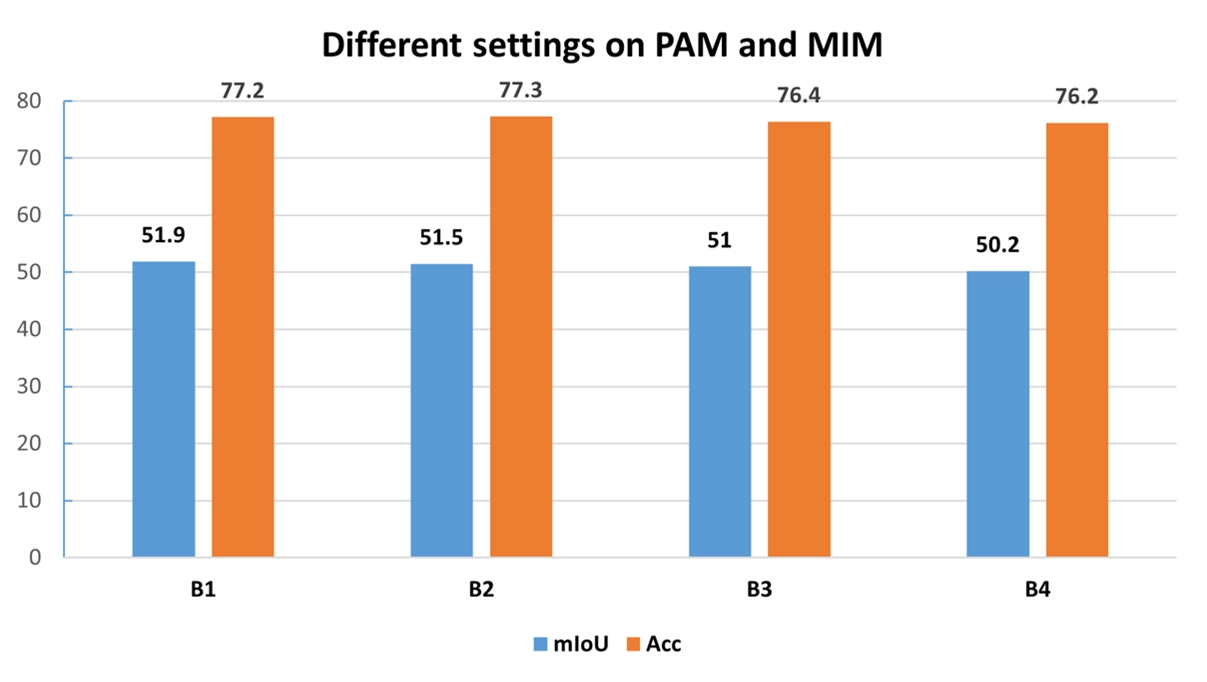}
\caption{Ablation Study on PAM and MIM. When set to B1, the best segmentation result is 51.9\%\label{fig:Ablation}}
\end{figure}

We conducted ablation experiments comparing PAM and MIM on the NYUv2 dataset as show in Fig. \ref{fig:Ablation}.
Specifically, the RGB feature and depth feature input PAM to obtain $\tilde{\bm F}_{RGB}^{n}$ and $\tilde{\bm F}_{Dep}^{n}$. Given the modality differences, we addressed the parameter-sharing issue in PAM. Moreover, considering the impact of network depth on information interaction, we applied MIM in both Layer 3 and Layer 4 of the encoder. Fig. \ref{fig:Ablation} presents the results of ablation studies on PAM and MIM using different configurations (B1-B4) on the NYUv2 dataset: B1 (PAM without shared parameters and MIM used on ResNetLayer4), B2 (PAM with shared parameters and MIM used on ResNetLayer4), B3 (PAM without shared parameters and MIM used on ResNetLayer3 and ResNetLayer4), B4 (PAM with shared parameters and MIM used on ResNetLayer3 and ResNetLayer4). Achieving the best results involves using PAM without shared parameters in ResNetLayer1-3 and MIM only in the last layer of the encoder, resulting in the highest mIoU of 51.9\%.
\subsection{Ablation Study on NYUv2 and SUN-RGBD}
To investigate the impact of different modules on segmentation performance, we conducted ablation experiments on NYUv2 and SUN-RGBD datasets, as depicted in Table \ref{tab:3}. '\usym{2713}' indicates the usage of a particular module, while '\usym{2715}' means not using the module. For instance, our PAM module exhibited a superiority of 1.5\% and 0.9\% over the baseline concerning mIoU and Acc indicators. Similarly, our MIM module demonstrated a superiority of 3.7\% and 1.9\% over the baseline regarding mIoU and Acc indicators. The result suggests that each proposed module can independently enhance segmentation accuracy.Our module surpasses the baseline in fusing cross-modal features, yielding superior results on both datasets. Using both PAM and MIM modules, we achieved the highest mIoU of 51.9\% on the NYUv2 dataset and the highest mIoU of 48.8\% on the SUN RGB-D dataset. The result highlights that our two designed modules can be collectively optimized to enhance segmentation accuracy.

\begin{table}[htb]
  \centering
  \caption{Ablation studies on NYUDv2 and SUN-RGBD dataset for PAM and MIM}
    \begin{tabular}{c||cccccc}
    \hline
    \hline
    \multirow{4}[4]{*}{Method} & \multirow{4}[4]{*}{\textbf{PAM}} & \multirow{4}[4]{*}{\textbf{MIM}} & \multicolumn{2}{c}{\multirow{2}[2]{*}{NYUv-2}} & \multicolumn{2}{c}{\multirow{2}[2]{*}{SUN-RGBD}} \bigstrut[t]\\
          &       &       & \multicolumn{2}{c}{} & \multicolumn{2}{c}{} \bigstrut[b]\\
\cline{4-7}          &       &       & \multirow{2}[2]{*}{mIoU(\%)} & \multirow{2}[2]{*}{Acc(\%)} & \multirow{2}[2]{*}{mIoU(\%)} & \multirow{2}[2]{*}{Acc(\%)} \bigstrut[t]\\
          &       &       &       &       &       &  \bigstrut[b]\\
    \hline
    \multirow{2}[2]{*}{Baseline} & \multirow{2}[2]{*}{\usym{2715}} & \multirow{2}[2]{*}{\usym{2715}} & \multirow{2}[2]{*}{47.4} & \multirow{2}[2]{*}{75.1} & \multirow{2}[2]{*}{45.5} & \multirow{2}[2]{*}{81.1} \bigstrut[t]\\
          &       &       &       &       &       &  \bigstrut[b]\\
    \hline
    \multirow{5}[2]{*}{Ours} & \usym{2713}     &\usym{2715}       & 48.9  & 76.0    & 47.9  & 81.3 \bigstrut[t]\\
          &       &       &       &       &       &  \\
          &\usym{2715}       & \usym{2713}     & 51.1  & 77.0    & 48.3  & 81.5 \\
          &       &       &       &       &       &  \\
          & \usym{2713}     & \usym{2713}     & \textbf{51.9} & 77.2  & \textbf{48.8} & 82.3 \bigstrut[b]\\
    \hline
    \end{tabular}%
  \label{tab:3}%
\end{table}%

\section{Conclusions}

In this paper, we tackle a fundamental challenge in RGB-D semantic segmentation—efficiently fusing features from two distinct modes. We designed an innovative Multi-modal Interaction and Pooling Attention network, which uses a small and flexible PAM module in the shallow layer of the network to enhance the feature extraction capability of the network and uses a MIM module in the last layer of the network to integrate RGB features and depth features effectively. We use the complementary information between RGB and depth mode to improve the accuracy of semantic segmentation in indoor scenes. In future work, we will extend our method to enhance its generalization ability in RGB-D semantic segmentation. Furthermore, we anticipate performance improvements by integrating tasks like depth estimation into the existing framework, facilitating collaborative network interactions.
{\textbf{limitation.}} Our method's effectiveness has been exclusively validated on CNN networks, but we haven't verified other network architectures, such as Transformer. In addition, during the segmentation verification on the test set, the requirement to input both RGB and depth images limits the network's generalization ability. Consequently, the network may not achieve optimal segmentation results for datasets lacking depth information.

\section*{Acknowledgments}

All sources of funding of the study must be disclosed.

\section*{Conflict of interest}

The authors declare there is no conflict of interest.

\bibliographystyle{elsarticle-num} 
\bibliography{ref}



\end{document}